\documentclass[review]{elsarticle}

\usepackage{hyperref}

\journal{Journal of \LaTeX\ Templates}
\usepackage{geometry}
\usepackage{graphicx} 
\usepackage{indentfirst}
\usepackage{amsmath}
\usepackage{booktabs}
\usepackage{multirow}
\usepackage{amsmath,bm}
\begin{document}

\begin{frontmatter}

\title{Medical image super-resolution method based on dense blended attention network}

\author[mymainaddress]{Kewen Liu}
\author[mymainaddress]{Yuan Ma}
\author[mysecondaryaddress]{Hongxia Xiong\corref{mycorrespondingauthor}}
\cortext[mycorrespondingauthor]{Corresponding author}
\ead{xionghongxia@whut.edu.cn}
\author[mythirdaddress]{Zejun Yan}
\author[myfourtharyaddress]{Zhijun Zhou}
\author[mymainaddress]{Panpan Fang}
\author[myfifthaddress]{Chaoyang Liu}

\address[mymainaddress]{School of Information Engineering, Wuhan University of Technology, Wuhan 430070, China}
\address[mysecondaryaddress]{School of Civil Engineering \& Architecture, Wuhan University of Technology, Wuhan 430070, China}

\address[mythirdaddress]{Department of Urology, Ningbo First Hospital, Key Laboratory of Translational Medicine of Urological Diseases in Ningbo, Ningbo 315010, China}
\address[myfourtharyaddress]{Department of Urology, the First People’s Hospital of Tianmen, Tianmen 431700, China}
\address[myfifthaddress]{State Key Laboratory of Magnetic Resonance and Atomic Molecular Physics, Wuhan Institute of Physics and Mathematics, Chinese Academy of Sciences, Wuhan 430071, China}

\begin{abstract}
In order to address the issue that medical image would suffer from severe blurring caused by the lack of high-frequency details in the process of image super-resolution reconstruction, a novel medical image super-resolution method based on dense neural network and blended attention mechanism is proposed. The proposed method adds blended attention blocks to dense neural network(DenseNet), so that the neural network can concentrate more attention to the regions and channels with sufficient high-frequency details. Batch normalization layers are removed to avoid loss of high-frequency texture details. Output high resolution medical images are obtained using deconvolutional layers at the very end of the network as up-sampling operators. Experimental results show that the proposed method has an improvement of 0.05db to 11.25dB and 0.6\% to 14.04\% on the peak signal-to-noise ratio(PSNR) metric and structural similarity (SSIM) metric, respectively, compared with the mainstream image super-resolution methods. This work provides a new idea for theoretical studies of medical image super-resolution reconstruction.
\end{abstract}


\end{frontmatter}

\section{Introduction}
\setlength{\parindent}{2em}
Image super-resolution(SR) refers reconstructing corresponding high-resolution(HR) image according to its low-resolution(LR) counterpart\cite{1}. According to the number of input frames, SR can be classified into single-image super-resolution(SISR) and multi-image super-resolution(MISR). This work focuses on SISR. Compared with methods which rely on hardware improvements, using SR algorithms for fast blurry images super-resolution reconstruction is characterized by stable universality, high efficiency and low cost\cite{2}\cite{3}.\\
\indent With the rapid development of artificial intelligence technology, image super-resolution technology has been widely used in medical image, which is one of the research hotspots in the field of medical image processing, computer-aided diagnosis and other fields\cite{4}. Doctors are able to see the biological structures and early lesions more clearly with the high-resolution medical images obtained by deploying super-resolution algorithms, which is of considerable beneficial of diagnosing and treating diseases\cite{5}\cite{6}\cite{7}\cite{8}.\\
\indent According to different principles, image super-resolution methods can be divided into three categories: interpolation-based, model-based and learning-based methods\cite{9}. Interpolation-based methods such as bicubic interpolation have the characteristic of high computational efficiency, but they are easy to lose high-frequency texture details. Model-based methods such as the maximum a posteriori probability(MAP) method which use the prior information to constrain the solution space\cite{10}. The performance is improved compared to interpolation-based methods, however, there is little prior information that can be effectively utilized when size of the input image is small, causing poor performance. \\
\indent Learning-based methods can be divided into compressed sensing-based methods\cite{11} and deep learning-based methods\cite{12}. Compressed sensing is a technique to acquire and reconstruct signals efficiently, by finding solutions to underdetermined linear systems. This is based on the principle that, through optimization, the signal sparsity can be exploited to recover it from far fewer samples than required by the Nyquist–Shannon sampling theorem\cite{11}. Yang et al. first introduce compressed sensing into the field of image super-resolution, and propose a method based on sparse representation which simultaneously learns the high resolution dictionary and the low resolution dictionary so that the high resolution image blocks and their corresponding low resolution image blocks have the same sparse coding in their own dictionaries\cite{11}. Li et al. introduce a low-dose PET images super-resolution method based on sparse representation whose experimental results show that considerable results have been obtained\cite{13}. Methods based on sparse representation can better preserve edge textures but are difficult to learn higher-level abstract features compared to interpolation-based and model-based methods, furthermore, they are incapable when the scaling ratio of super-resolution is large. \\
\indent Dong et al. first propose a deep learning-based method, super-resolution using convolutional neural network (SRCNN)\cite{12}, which is divided into three stages, namely feature extraction, nonlinear mapping, and reconstruction, to actualize end-to-end learning. Convolutional neural network-based methods no longer explicitly learn an external dictionary, but implicitly learn the convolution kernel parameters of the middle layers of the network, which have better generalization and expression ability than traditional methods. On the basis of SRCNN, Kim et al. propose VDSR(Very Deep Super-Resolution Algorithm) and draw the conclusion that the deeper the network, the better the performance\cite{14}. Compared with SRCNN, VDSR deepens the network layers, adds a skip connection to learn the residual between input and output images which is beneficial of improving the gradient vanishing and network degradation problems. Both SRCNN and VDSR need to be upsampled via bicubic interpolation before they are fed into the network which means that the convolution operation is performed on a large-sized image space, resulting in complicated calculation and low efficiency. In order to address the aforementioned issue, Dong et al. propose FSRCNN (Fast SRCNN)\cite{15}. FSRCNN performs convolution directly on the low-resolution image space whose size is smaller than that via bicubic interpolation, and performs deconvolution at the top of the network to obtain the final high-resolution image. Compared with SRCNN and VDSR, FSRCNN achieves significant improvements on efficiency and reconstruction results. Residual net(ResNet) proposed by He et al.\cite{16} form a deep network by stacking multiple residual blocks, which alleviates the gradient vanishing and network degradation problems caused by network’s depth deepening. Ledig et al. propose SRResNet (Super-Resolution Residual Network) based on ResNet\cite{17}. SRResNet introduces abundant global and local skip connections, so that the majority of low-frequency texture contents can be directly transmitted to the top layer of the network by the skip connections, with which bring the advantages of alleviating gradient vanishing and enhancing feature propagation. Dense neural network(DenseNet) proposed by Huang et al.\cite{18} adds dense skip connections, so that the output feature maps of any layer can be transmitted to subsequent layers through the dense skip connections, as part of the input of subsequent layers. This structure can fully multiplex features from different stages and different scales, and achieves better performance with less parameters and lower calculation costs than traditional residual network (ResNet). Tong et al. introduce the dense neural network into image super-resolution field and propose SRDenseNet (Super-Resolution Dense Network)\cite{19}, which achieves considerable results.\\
\indent Attention mechanism refers to that neural networks are capable of focusing on specific channels or specific regions. According to the differences of concerns, attention mechanism can be divided into spatial attention mechanism and channel attention mechanism\cite{20}. The cascaded channel-space attention mechanism proposed by Chen et al.\cite{21} cascades the channel attention block and the spatial attention block together, and learns corresponding descriptors for different channels and different regions in stages, assigning different channels and different regions different weights, forcing the neural network concentrate more attention on the channels and regions with sufficient high-frequency details. Another channel attention structure proposed by Hu et al.\cite{22} can adaptively assign different weights to different channels, enhance the channels with abundant high-frequency details, and suppress the channels with plentiful redundant low-frequency texture contents. The structure is capable of accelerating network convergence and further improves network performance. Zhao et al. indicate that the L2 norm-based loss function is a differentiable convex function\cite{23} and obtained images by models using L2 norm-based loss function are of favorable quality and can gain high peak signal-to-noise ratio (PSNR) metric results.\\
\indent In order to address the issue that medical image would suffer from severe blurring caused by the lack of high-frequency details in the process of image super-resolution reconstruction, a novel medical image super-resolution method based on dense neural network and blended attention mechanism is proposed. Our main contributions can be summarized as follows:

\begin{enumerate}[1.]
\item A new blended attention mechanism block is proposed. The proposed attention block learns the corresponding descriptors for input feature maps, multiplies the learned descriptors using $Hadamard$ product with the original input, and simultaneously assigns different weights to different channels and different regions to enhance the channels and regions with sufficient high-frequency details and suppress the channels and regions with abundant low-frequency texture contents. The proposed attention block is capable of enhancing feature representation capabilities, allowing the network to simultaneously focus on channels and regions that are with sufficient high-frequency details. Compared with phased learning, the proposed block has fewer network parameters and gains higher efficiency.
\item An image super-resolution method based on dense blended attention network is proposed. The basic network structure is based on dense neural network, dense skip connections are added between the basic unit of the network and inside each basic unit, with which can fully multiplex features from different stages and different scales. Original batch normalization layers in DenseNet are removed to avoid loss of high-frequency texture details. At the end of each basic unit, the proposed blended attention mechanism block is added, so that the neural network can concentrate attention more on the channels and regions with sufficient high-frequency details, which can accelerate network convergence and further improve the performance of the network.
\item The proposed method was applied to the super-resolution reconstruction of blurry prostate cancer MRI images. Experimental results show that the high-resolution MRI images obtained by performing super-resolution with scaling ratio 2, 3, and 4 can still preserve favorable image sharpness and texture details. In addition to applying to MRI images, the proposed method can also be applied to X-ray computed tomography(CT) images, X-ray(X-ray) images, and positron emission computed tomography(PET) images super-resolution through transferring network structures, indicating favorable universality of the proposed method.
\end{enumerate}

\section{Related theory}
\subsection{Dense neural network}
\indent The basic structure of a dense neural network is shown in Figure 1\cite{18}.

{\centering\includegraphics[height=3cm]{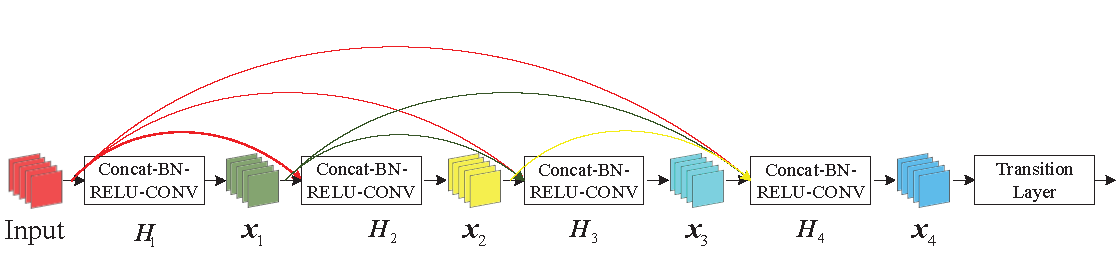}

}
{\centering Fig.1 Basic structure of primitive dense neural network

}
\indent The dense neural network is characterized by adding densely connected skip connections, fully multiplex features from different stages and different scales, and achieving better performance with less parameters and lower calculation costs than traditional residual network (ResNet). Besides, the dense skip connections allow the gradient information to be directly transmitted to any layer of the network during back propagation, greatly alleviating the problems of gradient vanishing and network degradation.\\
\indent The essential difference between DenseNet and ResNet is that the input of any layer in DenseNet is derived from the output of all previous layers. Assuming the DenseNet has $L$ layers, the output of ${l_{th}}$ layer of DenseNet $\bm{x_l}$ is:
\begin{equation}
\bm{x_l} = \bm{{H_l}}{\rm{(}}{\mathop{\rm concat}\nolimits} {\rm{(}}\bm{{x_0},{x_1}{\rm{,}}{x_2} \cdot  \cdot  \cdot {x_{l - 1}}}{\rm{)),  }}\bm{{x_0},{x_1}{\rm{,}}{x_2}} \cdot  \cdot  \cdot \bm{{x_{l - 1}},{x_l}} \in {\bm{R}^{H*W*C}}
\end{equation}
where $concat$ is the concatenation operation on the channel dimension. Equation 1 indicates that, the input of ${l_{th}}$ layer is derived not only from the output of ${(l - 1)_{th}}$ layer, but also from the output of all previous layers. Concatenate $\bm{{x_0},{x_1}{\rm{,}}{x_2} \cdot  \cdot  \cdot {x_{l - 1}}}$ in channel dimension, and take the concatenated feature maps as the input of ${l_{th}}$ nonlinear transforming operator $\bm{H_l}$. However, the output of ${l_{th}}$ layer of ResNet is:
\begin{equation}
\bm{x_l} = \bm{H_l}(\bm{x_{l - 1}}) + \bm{x_{l - 1}},\bm{x_{l - 1}}{\rm{,  }}\bm{x_l} \in {\bm{R}^{H*W*C}}
\end{equation}
Namely, the output of ${l_{th}}$ layer is only the pixel-wise sum of the output of ${(l - 1)_{th}}$ layer and the nonlinear transformation of the output of ${(l - 1)_{th}}$ layer. \\
\indent In view of the fact that DenseNet is capable of fully multiplexing features from different stages and different scales, and achieving better performance with less parameters and lower calculation costs than traditional residual network (ResNet). The proposed method in this paper is based on dense neural network.

\subsection{Blend attention mechanism}
\indent Adding a blended attention mechanism block in the network forces the neural network concentrates more attention on the channels and regions with sufficient high-frequency details, which is capable of accelerating network convergence and further improves network performance. The proposed attention block cascades two convolutional layers and two activation layers, simultaneously learns corresponding descriptors for different channels and different regions, and assigns different channels and different regions different weights. Compared with phased learning, the proposed block has fewer network parameters and gains higher efficiency.

{\centering\includegraphics[height=2cm]{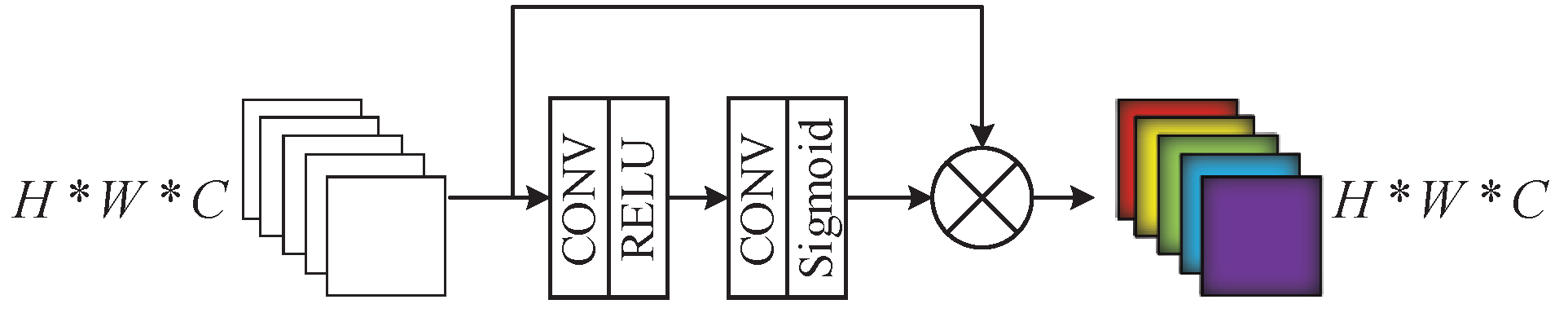}

}
{\centering Fig.2 Blended attention block

}
\indent The proposed blended attention mechanism block is shown in Figure 2. The dimensions of the input and output feature maps are both $H*W*C$, $conv$ represents convolutional operations, $RELU$ and $Sigmoid$ are two different activation functions\cite{24}, symbol$ \otimes $ is $Hadamard$ product\cite{25}. Taking feature maps with dimensions $H*W*C$ as input, after two cascaded convolutions and activations as the figure 2 shows, corresponding descriptors $\tau $ will be obtained:
\begin{equation}
\bm{\tau}  = f{\rm{(}}\bm{W_2}\delta {\rm{(}}\bm{W_1}\bm{x}{\rm{)),}}\tau  \in {\bm{R}^{H*W*C}}
\end{equation}
where $\bm{W_1},\bm{W_2}$ represent the parameters of the first and the second convolutional layer, respectively. $\delta$ represents $RELU$ activation function while $f$ represents $Sigmoid$ activation function. The first convolutional layer performs channel down-scaling with reduction ratio 16, after that, feature maps with dimensions $H*W*C/16$ can be obtained. The obtained feature maps are then increased by ratio 16. After dimensions down-scaling and up-scaling by two cascaded convolutional layers and two activation layers, $C$ corresponding descriptive matrices namely descriptors $\tau $ for different channelswhere $i = 0,1,2...C$  are learned. Sparser descriptive matrices are adaptively assigned to channels that contain more low-frequency texture contents, this enables the neural network concentrate more attention on the channels and regions with sufficient high-frequency details. The dimension of each descriptive matrix ${\tau _i}$ is $H*W$, corresponding to each element in ${i_{th}}$ channel of the original input feature maps. After two convolutions and two activations, the channels with abundant high-frequency details are enhanced, and the channels with plentiful redundant low-frequency texture contents are suppressed. Multiply the learned descriptor ${\tau _i}$ using $hadamard$ product with the ${i_{th}}$ channel to force the network focus on the regions with sufficient high-frequency details in the ${i_{th}}$ channel. In summary, feature representation through blended attention mechanism block can be obtained by multiplying the learned descriptor $\bm{\tau} $ and the original input.\\
\indent In view of the fact that blended attention mechanism is capable of accelerating network convergence and further improving network performance. The proposed method in this paper introduces blended attention mechanism on dense neural network.

\section{Methods}
\subsection{Network structure}
\indent The basic unit of the proposed dense blended attention network in this paper is shown in Figure 3a, where C represents channel concatenating, $RELU$ and $Sigmoid$ are two different activation functions. Each basic unit consists of eight cascaded convolutional layers, activation layers, and proposed blended attention blocks. Densely connected skip connections are added inside each basic unit, with which can fully multiplex features from different stages and different scales. Specifically, eight convolutional and activation operations are performed on the feature maps. The size of each convolution kernel is set to 16$\times$3$\times$3, namely 16 kernels whose size is 3$\times $3. RELU is used as activation function, convolutional step size is set to 1, zero padding is used to keep the size of feature maps same\cite{26}. Features from different stages and different scales are multiplexed due to dense skip connections. The number of channels of the convolution kernel increases linearly as the network deepens. In each basic unit, the number of channels of the first convolutional kernel is set to 16, and the number increases in each subsequent layer by 16 than the previous layer. After cascaded eight convolutions and activations, the feature maps are fed into the proposed blended attention block, and then output to the subsequent basic unit of the dense neural network to extract deeper features representations.

{\centering\includegraphics[height=5cm]{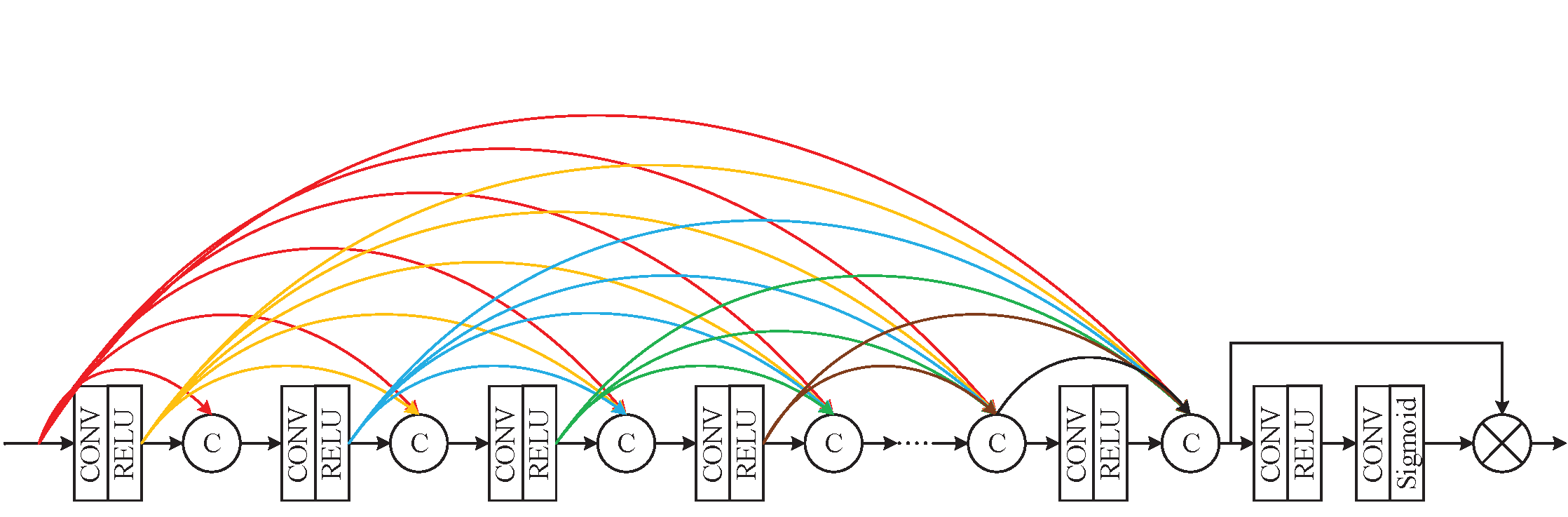}

}
{\centering (a) Basic unit

}
{\centering\includegraphics[height=5cm]{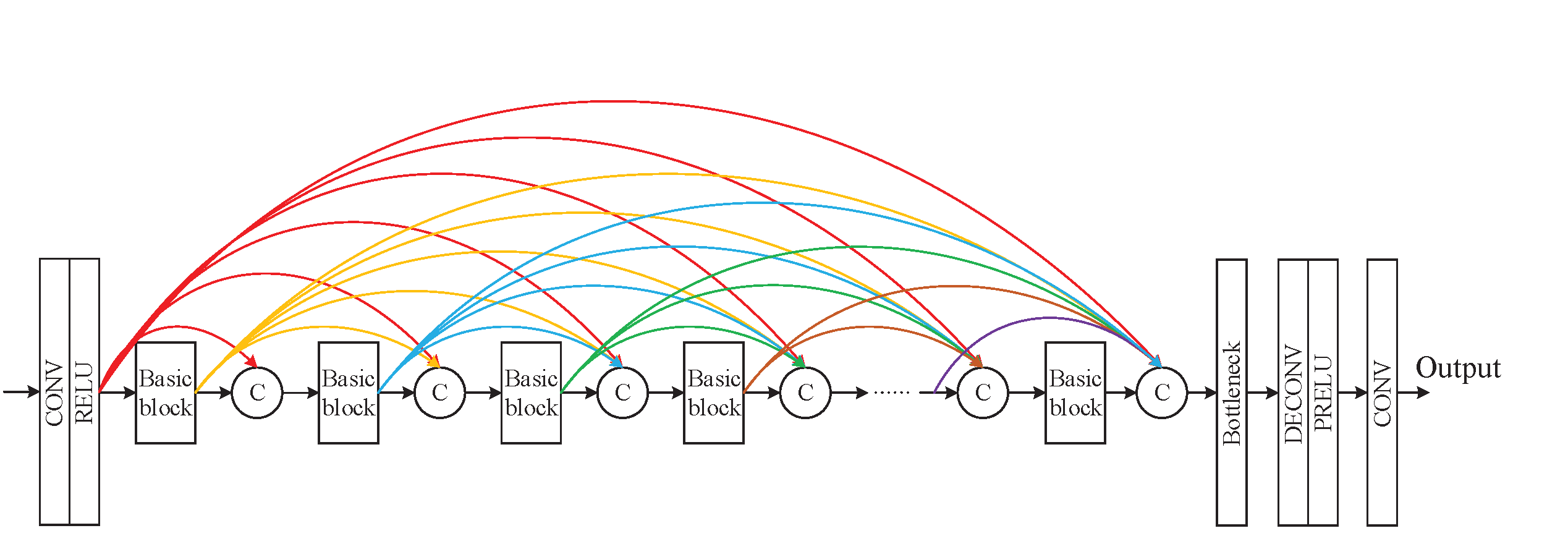}

}
{\centering (b) Proposed network structure

}
{\centering Fig.3 Proposed basic unit and network structure

}
The overall schematic of the network is shown in Figure 3b, where $Bottleneck$ represents the bottleneck layer\cite{17}], $Deconv$ represents the deconvolutional layer\cite{15}, $RELU,PRELU$ are two different activation functions. The whole network can be divided into three parts, namely feature extraction, nonlinear mapping, and reconstruction. The feature extraction part consists of concatenated convolution and activation layers. The convolutional kernel size is set to 128$\times $3$\times $3$\times $3, that is, 128 convolution kernels with a size of 3$\times $3, and the number of channels per kernel is 3. The nonlinear mapping part consists of eight cascaded basic units. Dense skip connections are added between the basic unit of the network, fully multiplexing features from different stages and different scales, greatly alleviating the problems of gradient vanishing and network degradation. The outputs of the eight basic units are concatenated together, and then fed into the subsequent bottleneck layer to perform channel-downscaling to reduce network parameters. After that, the final output high-resolution images are amplified by the cascaded deconvolution layers and activation layers.

\subsection{Loss function}
Inspired by the literature\cite{23}, we use the L2 norm-based loss to quantify the similarity between high-resolution images and high-resolution images obtained by super-resolution to guide network learning. The expression of the used loss function is:
\begin{equation}
{L_2} = \frac{1}{{n*H*W*C}}\sum\limits_{v = 1}^n {\sum\limits_i^W {\sum\limits_j^H {\sum\limits_k^C {{{(I_{v,i,j,k}^{HR} - I_{v,i,j,k}^{SR})}^2}} } } }
\end{equation}
where ${I^{HR}}$ represents real high resolution images, ${I^{SR}}$ represents the high-resolution images obtained by super-resolution, $H,W,C$ are the size and channel number of the input image, $n$ represents the number of mini-batch learning, $I_{v,i,j,k}^{}$ is the pixel value of position $(i,j)$ in the ${k_{th}}$ channel of the ${v_{th}}$ input image. The goal of the network in the training phase is to minimize the loss function ${L_2}$. The smaller the ${L_2}$ loss, the smaller the difference between the high-resolution image obtained by deploying super-resolution and the real high-resolution image, the better the super-resolution performance and the higher the precision. 

\subsection{Training details}
Training HR images are cropped to sub-images of size 96$\times $96 as preprocessing procedure. LR images with scaling ratio 2, 3 and 4 are obtained by down-sampling HR images using the MATLAB bicubic kernel function. Data augmentation is performed on the training images, which are randomly rotated by ${90^\circ }$,${180^\circ }$,${270^\circ }$ and flipped horizontally to obtain more training data. The mini-batch size is set to 16 due to hardware resource limitations. For optimization, the proposed method is optimized by ADAM optimizer with ${\beta _1}{\rm{ = }}0.9$, ${\beta _2}{\rm{ = }}0.999$, and ${\epsilon}{\rm{ = }}{10^{ - 8}}$\cite{27}. The initial learning rate is set to ${10^{ - 4}}$ and then decreases to half when there is no precision improvement in 10 consecutive epochs to achieve optimal results.

\section{Simulation}
\subsection{Experimental parameters settings}
Training images used in this paper were prostate MRI public datasets of the University of Medicine, Nijmegen, Netherlands, which contains T2W, PD-W, DCE and DW images, all of which were derived from two different Siemens 3T MR scanners, the MAGNETOM Trio and Skyra\cite{28}. T2-weighted images were acquired using a turbo spin echo sequence and had a resolution of around 0.5 mm in plane and a slice thickness of 3.6 mm. 400 T2W high-resolution prostate MRI images with sufficient details whose size is 384$\times $384 were selected as training set, and 100 T2W prostate MRI images were selected as testing set. Two traditional super-resolution methods and four representative deep learning-based methods were selected as comparative experiments. The traditional methods selected were bilinear interpolation (Bilinear) and bicubic interpolation (Bicubic). The selected deep learning methods were SRCNN\cite{12}, VDSR\cite{14}, SRResNet\cite{7} and SRDenseNet\cite{18}. For fair comparison, each method was tested under the same hardware environment. The hardware environmental parameters of the experiment are 
shown in Table 1.

{\centering Table 1. Experimental hardware environment parameters

}

{\centering
\makeatletter\def\@captype{table}\makeatother
\begin{tabular}{cc}
\toprule[0.5pt]
Hardware Configuration & Parameters  \\
\midrule[0.5pt]                        
CPU                    & Intel(R) Xeon(R) E3 1231V3 @ 3.4GHz \\
RAM                    & 16G                                 \\
GPU                    & 1070Ti                              \\
GPU Memory             & 8G                                  \\
Development Framework  & Pytorch1.0.1    \\
\bottomrule[0.5pt]                   
\end{tabular}

}

\subsection{Quantitative evaluation}
The metrics widely used to evaluate the image super-resolution performance are Peak Signal-to-Noise Ratio (PSNR)\cite{29} and Structural Similarity (SSIM)\cite{30}. PSNR and SSIM were chosen for quantitative evaluation in this paper. In addition, the real-time performance of the super-resolution algorithm is also a vital indicator. The time required to complete the super-resolution of single-frame image is also used as one of the quantitative evaluation metrics.
For evaluation, inspired by the literature\cite{31}, this paper converted images from RGB color space to YCbCr color space\cite{32}. All the metric results reported in this paper were computed on the y-channel after removing a 4-pixel border. The formula for calculating PSNR is:
\begin{equation}
PSNR = 10{*}\lg \frac{{{{255}^2}}}{{\frac{1}{{W*H}}\sum\limits_{i = 1}^W {\sum\limits_{j = 1}^H {{\rm{(}}{I_{i,j}}^{HR} - {I_{i,j}}^{SR}{\rm{)}}} } }}
\end{equation}
where $W,H$ represents the image size, ${I^{HR}}$ represents real high-resolution images, ${I^{SR}} $ represents the high-resolution images obtained by super-resolution, $I_{i,j}^{}$ represents the pixel value of position $(i,j)$. The better the PSNR results, the better the image quality.
The formula for calculating SSIM is:
\begin{equation}
SSIM(x,y) = \frac{{2{u_x}{u_y} + {C_1}}}{{{u_x}^2 + {u_y}^2 + {C_1}}}*\frac{{2{\sigma _x}{\sigma _y} + {C_2}}}{{{\sigma _x}^2 + {\sigma _y}^2 + {C_2}}}*\frac{{{\sigma _{xy}} + {C_2}}}{{{\sigma _x}{\sigma _y} + {C_3}}}
\end{equation}
where $u,\sigma $ represents the pixel mean and variance of two images for comparison, ${C_1},{C_2},{C_3}$ represents the constants to prevent the denominator from being zero. The range of SSIM is ${\rm{[0,1]}}$, the closer the value is to 1, the more similar the two images.
The PSNR, SSIM, and time-consuming metric results obtained by super-resolution of the testing set were averaged. The results are shown in Table 2.
Taking the quantitative evaluation results of scaling ratio 2 as an example, the proposed method has 11.247dB and 14.036\% improvement in peak signal-to-noise ratio (PSNR) and structural similarity (SSIM), respectively, compared with bicubic interpolation. Compared with SRCNN, there is 5.523dB and 2.738\% improvement, compared with VDSR with 0.962dB and 0.331\% improvement, compared with SRResNet with 0.640dB and 0.225\% improvement, compared with SRRenseNet with 0.196dB and 0.6\% improvement. When it came to time-consuming, bilinear interpolation was the fastest, time consumption of SRCNN, VDSR, SRResNet, SRDenseNet and the proposed method were higher than that of bilinear interpolation, however, they are all within 0.5s which have favorable real-time performance.\\
\indent It can be seen from the quantitative evaluation results of scaling ratio of 2, 3, 4 that performance of the proposed method is superior to the comparative methods. Combining the performance of each method of different scaling ratio, the conclusion that the of the proposed method is effective and superior can be drawn.

{\centering Table 2.  Quantitative evaluation results of each method

}

{
\centering
\makeatletter\def\@captype{table}\makeatother
\begin{tabular}{cccccccccc}
\toprule[0.5pt]
\multirow{2}{*}{Method}
&\multicolumn{3}{c}{Scaling ratio: 2} &\multicolumn{3}{c}{Scaling ratio: 3}&\multicolumn{3}{c}{Scaling ratio: 4}\\
& PSNR/dB & SSIM & TIME/s & PSNR/dB & SSIM & TIME/s & PSNR/dB & SSIM & TIME/s \\
\midrule[0.5pt]
\begin{tabular}[c]{@{}c@{}}Bilinear \\Bicubic  \\SRCNN\end{tabular}& 
\begin{tabular}[c]{@{}c@{}}24.457   \\25.5926  \\30.1813 \end{tabular} & 
\begin{tabular}[c]{@{}c@{}}0.79361  \\0.85839  \\0.90659\end{tabular} & 
\begin{tabular}[c]{@{}c@{}}0.0357   \\0.0392   \\0.3134\end{tabular} & 
\begin{tabular}[c]{@{}c@{}}23.5452  \\24.0203  \\24.617\end{tabular} & 
\begin{tabular}[c]{@{}c@{}}0.71903  \\0.75901  \\0.81318\end{tabular} & 
\begin{tabular}[c]{@{}c@{}}0.0344   \\0.0365   \\0.3039\end{tabular} & 
\begin{tabular}[c]{@{}c@{}}22.6134  \\23.1547   \\23.462\end{tabular} & 
\begin{tabular}[c]{@{}c@{}}0.63552  \\0.68166   \\0.73439\end{tabular} & 
\begin{tabular}[c]{@{}c@{}}0.0325    \\0.0359   \\0.3180\end{tabular} \\
\begin{tabular}[c]{@{}c@{}}VDSR\\SRResNet\end{tabular}& 
\begin{tabular}[c]{@{}c@{}}34.7425\\35.0647\end{tabular}& 
\begin{tabular}[c]{@{}c@{}}0.93066\\0.93172\end{tabular}& 
\begin{tabular}[c]{@{}c@{}}0.3226\\0.3893\end{tabular}& 
\begin{tabular}[c]{@{}c@{}}28.0425\\31.0468\end{tabular}& 
\begin{tabular}[c]{@{}c@{}}0.82446\\0.86065\end{tabular}& 
\begin{tabular}[c]{@{}c@{}}0.3496\\0.3712\end{tabular}& 
\begin{tabular}[c]{@{}c@{}}24.1305\\29.116\end{tabular}& 
\begin{tabular}[c]{@{}c@{}}0.77406\\0.80235\end{tabular}& 
\begin{tabular}[c]{@{}c@{}}0.3454\\0.3766\end{tabular} \\
SRDenseNet & 35.5079 & 0.93337 & 0.4476 & 31.869 & 0.86593 & 0.4364 & 29.92 & 0.80955 & 0.4508 \\
Proposed method & 35.7043 & 0.93397 & 0.4623 & 31.9154 & 0.86704 & 0.4684 & 29.9721 & 0.81085 & 0.4701\\
\bottomrule[0.5pt]
\end{tabular}
}

\subsection{Qualitative evaluation}
\indent Four sets of images with sufficient texture details were selected from the testing set to show the performance of each super-resolution method. The four images selected were MRI images of the transverse transposition of the prostate datasets. The comparisons between the high-resolution images obtained by each method and the real high-resolution images (Ground Truth, GT) are shown in Figure 4, and the corresponding quantitative evaluation results are marked below.\\
\indent In overall, the images obtained by deep learning-based methods are clearer than that obtained by interpolation-based methods. SRCNN, VDSR, SRResNet, SRDenseNet and the proposed method in this paper achieved considerable performance, however, it can be noticed that more realistic texture details were produced by the proposed method in specific areas. Taking the ProstateX-0061 image shown in Figure 4 as an example, the result images obtained by bilinear and bicubic interpolation look blurry. The images obtained by SRCNN have certain improvements compared with bicubic interpolation, however, they are still blurry and over-smoothed that appear implausible. VDSR, SRResNet, and SRDenseNet achieved favorable sharpness, but blurry artifacts appear. Images obtained by the proposed method have gained better sharpness, uniform brightness, sufficient details and perception results, and are closest to the real high-resolution images.
\subsection{Comprehensive evaluation}
Combined with the quantitative and qualitative evaluation results, the method proposed in this paper has the characteristics of high precision. The quantitative results calculated by performing the proposed method on the testing set are higher than that of traditional interpolation-based method and the deep learning-based SRCNN, VDSR, SRResNet, and SRDenseNet methods. It can be noticed from the comparison of qualitative results of each super resolution method in Figure 4 that MRI images obtained by the proposed method have gained higher definition, better sharpness, uniform brightness, sufficient details and favorable perception results and are closet to the real high-resolution images. When it comes to the time-consuming, the proposed method improves performance with sacrifice of time, however, the consumed time is also within the acceptable range. In addition to applying to MRI images, the proposed method can also be applied to X-ray computed tomography(CT) images, X-ray (X-ray) images, and positron emission computed tomography (PET) images super-resolution through transferring network structures, indicating favorable universality of the proposed method.\\
\indent In summary, the proposed method is superior to traditional interpolation-based methods and deep learning-based SRCNN, VDSR, SRResNet and SRDenseNet methods.

\section{Conclusion}

In order to address the issue that medical image would suffer from severe blurring caused by the lack of high-frequency details in the process of image super-resolution reconstruction, a novel medical image super-resolution method was proposed. Experimental results showed the proposed method is superior to mainstream image super-resolution methods. This work provides a new idea for theoretical studies of medical image super-resolution reconstruction.

\section*{Acknowledgments}
This work was supported by the National Key R\&D Program of China (2018YFC0115000).

{\centering\includegraphics[height=23.9cm]{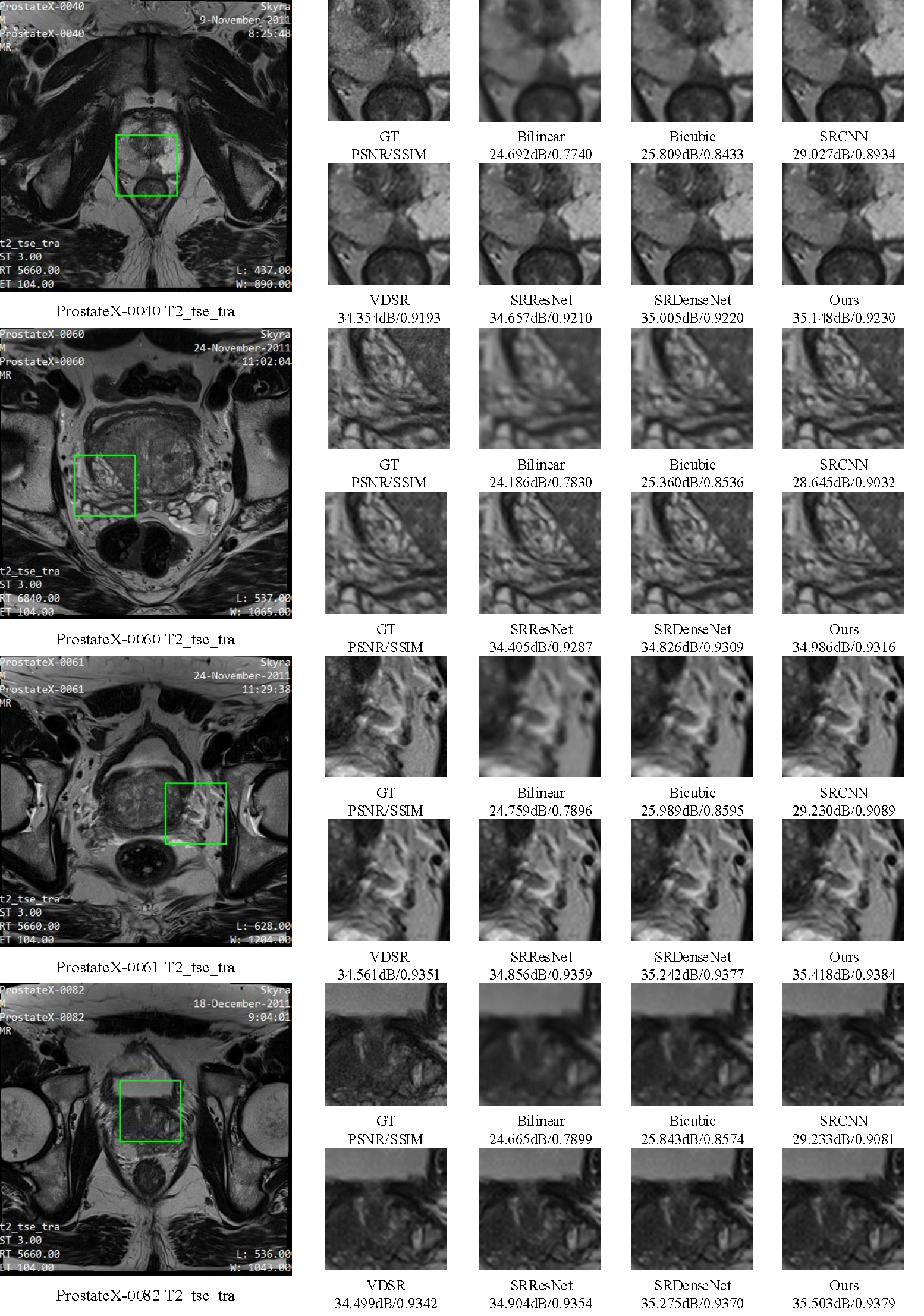}

}
{\centering Fig.4 Comparison of qualitative results of each super resolution method with scaling ratio 2

}

{
\footnotesize
\bibliography{elsarticle-template}
}

\end{document}